\documentclass{article}

\usepackage[T1]{fontenc}
\usepackage{amssymb,amsmath}
\usepackage{txfonts}
\usepackage{microtype}

\usepackage{graphicx}
\usepackage{subfigure} 

\usepackage{natbib}

\usepackage{algorithm}
\usepackage{algorithmic}

\usepackage{hyperref}
\usepackage{url}
\usepackage{mlp2017}
\DeclareMathOperator{\relu}{relu}
\DeclareMathOperator{\lrelu}{lrelu}
\DeclareMathOperator{\elu}{elu}
\DeclareMathOperator{\selu}{selu}

\usepackage{siunitx}
\usepackage{tabularx}

\def\studentNumber{\textbf{Dabal Pedamonti} }

\begin{document} 

\twocolumn[
\mlptitle{Comparison of non-linear activation functions for deep neural networks on MNIST classification task}

\centerline{\studentNumber}
\centerline{Department of Computer Science}
\centerline{University of Edinburgh}

\vskip 7mm
]

\begin{abstract} 
Activation functions play a key role in neural networks so it becomes fundamental to understand their advantages and disadvantages in order to achieve better performances. This paper will first introduce common types of non linear activation functions that are alternative to the well known sigmoid function and then evaluate their characteristics. Moreover deeper neural networks will be analysed because they positively influence the final performances compared to shallower networks. They also strictly depend on the weight initialisation hence the effect of drawing weights from Gaussian and uniform distribution will be analysed making particular attention on how the number of incoming and outgoing connection to a node influence the whole network.

\end{abstract} 

\section{Introduction}
\label{sec:intro}
The main activation function that was widely used is the Sigmoid function however, when the \textbf{Rectifier Linear Unit} (ReLU) \citep{relu00} was introduced, it soon became a better replacement for the Sigmoid function due to its positive impact on the different machine learning tasks.

After that, different variants of the ReLU activation function have been introduced and this experiment explores them and their impact on the MNIST (Modified National Institute of Standards and Technology) digits set. The MNIST \citep{mnist00} dataset has 50,000 training images, 10,000 validation images, and 10,000 test images, each showing a 28$\times$28 grey-scale pixel image of one of the 10 digits.

The second part of the experiment explores deeper networks, until the 8th hidden layers and their accuracies and errors are compared with the shallower structures. Moreover, different approaches to weight initialisation based on uniform distribution are investigated such as making the estimated variance of a unit independent of the number of incoming connections (fan\_in) or making the estimated variance of a unit\rq s gradient independent of the number of outgoing connections (fan\_out). Another approach corresponds to Glorot and Bengio\rq s \citep{pmlr-v9-glorot10a} combined initialisation and the effect of the gaussian distribution respect to the uniform is explored as well.

For the first part of the experiment a batch size of 50 and a number of 100 epochs are used while for the second part the batch size stays the same and the number of epochs are halved to 50.

\section{Activation functions}
\label{sec:actfn}
The whole experiment is based in particular on three activation functions which are variants of the ReLU.
The ReLU activation function has the following form:
\begin{equation}
  \relu(x) = \max(0, x) ,
\end{equation} 
which has the gradient:
\begin{equation}
  \frac{d}{dx} \relu(x) =
     \begin{cases} 
      0      & \quad \text{if } x \leq  0 \\
      1       & \quad \text{if } x > 0 .
    \end{cases} 
\end{equation}
It comes with the aim to solve the vanishing gradient and exploding gradient problems: while using Sigmoid and working on shallower layers doesn\rq t give any problem, some issues arise when the architecture becomes deeper because the derivative terms that are less than 1 will be multiplied each other many times that the values will become smaller and smaller until the gradient tends towards zero hence vanishing. On the other hand if the values are bigger than 1 then the opposite happens, with numbers being multiplied becoming bigger and bigger until they tend to infinity and explode the gradient. A good solution would be to keep the values to 1 so even when they are multiplied they don\rq t change. This is exactly what ReLU does: it has gradient 1 for positive inputs and 0 for negative ones. The fact that the gradient is zero might be seen as an issue at first but it actually helps to make the network sparse keeping the useful links.
Sparsity helps to keep the network less dense and decreases the computation, however once the gradient is zero the corresponding nodes don\rq t have any influence on the network anymore so they can\rq t contribute to the improvement of the learning. This is called dying ReLU problem and gave origin to a variant of the rectifier linear unit called Leaky ReLU (LReLU).

The \textbf{Leaky ReLU} \citep{lrelu00}, which is going to be the first activation function I am going to explore, has the following form: 
\begin{equation}
  \lrelu(x) =
  \begin{cases} 
      \alpha x      & \quad \text{if } x \leq  0 \\
      x       & \quad \text{if } x > 0 .
    \end{cases} 
\end{equation} 
which has the gradient:
\begin{equation}
  \frac{d}{dx} \lrelu(x) =
     \begin{cases} 
      \alpha      & \quad \text{if } x \leq  0 \\
      1       & \quad \text{if } x > 0 .
    \end{cases} 
\end{equation}
where $\alpha = 0.01$.

To overcome the dying problem, an alpha parameter has been added which is indeed the leak, so the gradient will be small but not zero. This reduces the sparsity but, on the other hand, makes the gradient more robust for optimisation since now the weight will be adjusted for those nodes that were not active with ReLU.

The second activation function to be examined is the \textbf{Exponential Linear Unit} (ELU) \citep{elu00} which is given as:
\begin{equation}
  \elu(x) =
  \begin{cases} 
      \alpha (\exp (x)-1)      & \quad \text{if } x \leq  0 \\
      x       & \quad \text{if } x > 0 .
    \end{cases} 
\end{equation} 
which has the gradient:
\begin{equation}
  \frac{d}{dx} \elu(x) =
     \begin{cases} 
      \elu(x) + \alpha      & \quad \text{if } x \leq  0 \\
      1       & \quad \text{if } x > 0 .
    \end{cases} 
\end{equation}
where $\alpha = 1$.

Since the learning can be made faster by centering the activations at zero, ELU uses the activation function in order to achieve mean zero, rather than using batch normalisation. What slows down the learning is the bias shift which is present in the ReLUs. Those have mean activation larger than zero and learning cause bias shift for the following layers. So ELU is a good alternative to ReLU since it decreases the bias shift by pushing the mean activation towards zero.

The final activation function to be analysed is the \textbf{Scaled exponential Linear Unit} (SELU) \citep{selu00}which is the following:
\begin{equation}
  \selu(x) = \lambda
  \begin{cases} 
      \alpha (\exp (x)-1)      & \quad \text{if } x \leq  0 \\
      x       & \quad \text{if } x > 0 .
    \end{cases} 
\end{equation} 
which has the gradient:
\begin{equation}
  \frac{d}{dx} \selu(x) = 
     \begin{cases} 
      \selu(x) + \lambda\alpha     & \quad \text{if } x \leq  0 \\
      \lambda       & \quad \text{if } x > 0 .
    \end{cases} 
\end{equation}
where $\alpha = 1.6733$ and $\lambda = 1.0507$.

SELU has self-normalising properties because the activations that are close to zero mean and unit variance, propagated through many network layers, will converge towards zero mean and unit variance. This, in particular, makes the learning highly robust and allows to train networks that have many layers.

\section{Experimental comparison of activation functions}
\label{sec:actexpts}
The behavior of Leaky ReLU, ELU, and SELU activation functions on the MNIST set are explored using 2 hidden layers, with 100 units per hidden layer. Those are then compared with the baseline systems using sigmoid units and ReLU units.
\begin{table}[h]
\vskip 1mm
\begin{center}
\begin{small}
\begin{sc}
\begin{tabular}{llllll}
\hline
\abovespace\belowspace
$\eta$ &Sigmoid & ReLU & LReLU & ELU & SELU \\
\hline
\abovespace
0.01 &$1.64e_-1$ & $9.56e_-3$ & $1.11e_-2$ & $2.13e_-2$ & $1.45e_-2$\\
0.05 &   $2.61e_-2$ & $4.03e_-4$ & $4.15e_-4$ & $5.91e_-4$ & $5.97e_-4$\\
0.1    & $5.57e_-3$ & $1.42e_-4$ & $1.44e_-4$ & $2.05e_-4$ & $2.07e_-4$\\
0.2    & $1.33e_-3$ & $5.12e_-5$ & $5.68e_-5$ & $7.78e_-5$ & $6.09e_-5$  \\
\hline
\end{tabular}
\end{sc}
\end{small}
\caption{Errors of different activation functions on the training set according to the corresponding learning rate $\eta$.}
\label{tab:table-err-training}
\end{center}
\vskip -1mm
\end{table}

\begin{table}[h]
\vskip 1mm
\begin{center}
\begin{small}
\begin{sc}
\begin{tabular}{llllll}
\hline
\abovespace\belowspace
$\eta$&Sigmoid & ReLU & LReLU & ELU & SELU \\
\hline
\abovespace
0.01 &   $1.65e_-1$ & $9.14e_-2$ & $9.64e_-2$ & $8.40e_-2$ & $9.64e_-2$ \\
0.05 &   $8.57e_-2$ & $1.11e_-1$ & $1.22e_-1$ & $1.21e_-1$ & $1.14e_-1$\\
0.1    & $8.33e_-2$ & $1.17e_-1$ & $1.20e_-1$ & $1.15e_-1$ & $1.25e_-1$\\
0.2    & $9.83e_-2$ & $1.25e_-1$ & $1.15e_-1$ & $1.20e_-1$ & $1.22e_-1$\\
\hline
\end{tabular}
\end{sc}
\end{small}
\caption{Error of different activation functions on the validation set according to the corresponding learning rate $\eta$.}
\label{tab:table-err-validation}
\end{center}
\vskip -1mm
\end{table}

\begin{table}[h!]
\vskip 1mm
\begin{center}
\begin{small}
\begin{sc}
\begin{tabular}{lccccr}
\hline
\abovespace\belowspace
$\eta$&Sigmoid & ReLU & LReLU & ELU & SELU \\
\hline
\abovespace
0.01 &   $0.953$ & $0.999$ & $0.999$ & $0.996$ & $0.998$ \\
0.05 &   $0.995$ & $1.000$ & $1.000$ & $1.000$ & $1.000$\\
0.1    & $1.000$ & $1.000$ & $1.000$ & $1.000$ & $1.000$\\
0.2    & $1.000$ & $1.000$ & $1.000$ & $1.000$ & $1.000$ \\
\hline
\end{tabular}
\end{sc}
\end{small}
\caption{Accuracies of different activation functions on the training set according to the corresponding learning rate $\eta$.}
\label{tab:table-acc-training}
\end{center}
\vskip -1mm
\end{table}
I have first started the analysis by running four times each one of the activation functions by changing the learning rate between 0.01, 0.05, 0.1 and 0.2.

\begin{table}[h!]
\vskip 1mm
\begin{center}
\begin{small}
\begin{sc}
\begin{tabular}{lccccr}
\hline
\abovespace\belowspace
$\eta$&Sigmoid & ReLU & LReLU & ELU & SELU \\
\hline
\abovespace
0.01 &   $0.955$ & $0.977$ & $0.976$ & $0.978$ & $0.974$ \\
0.05 &   $0.974$ & $0.979$ & $0.977$ & $0.978$ & $0.976$\\
0.1    & $0.979$ & $0.980$ & $0.979$ & $0.980$ & $0.978$\\
0.2    & $0.977$ & $0.979$ & $0.982$ & $0.981$ & $0.981$ \\
\hline
\end{tabular}
\end{sc}
\end{small}
\caption{Accuracies of different activation functions on the validation set according to the corresponding learning rate $\eta$.}
\label{tab:table-acc-validation}
\end{center}
\vskip -1mm
\end{table}

As it can be seen from the tables 1-4, when the learning rate is low the loss tends to be small and when it increases the loss increases as well. 
It can be tempting to choose a small value but I needed to pay attention to the accuracy which increases with the learning rate, hence it is necessary to find a good trade off. 

Comparing different values I could see that learning rate of 0.05 and 0.1 had the best results both in term of losses and accuracies. This trend is happening in all the five activation functions which confirms that the following experiments should be run tuning the learning rate between these values. Moreover, in this case I have picked the value of 0.1 because I gave slightly more importance to the accuracy rather than the error since MNIST is a balanced set and the final task is to classify the digits correctly, in fact the accuracy on validation set increases by 2-4\% when the learning rate goes from 0.05 to 0.1 as it can be seen on table 4.

As it can be seen from the graph on figure 1, when the learning rate increases, the validation set starts overfitting the training data earlier. So I didn\rq t pick high value of learning rate because they generalise less than smaller value even if they held a higher accuracy (Figure 2).

\begin{figure}[h]
\vskip 2mm
\begin{center}
\centerline{\includegraphics[width=\columnwidth]{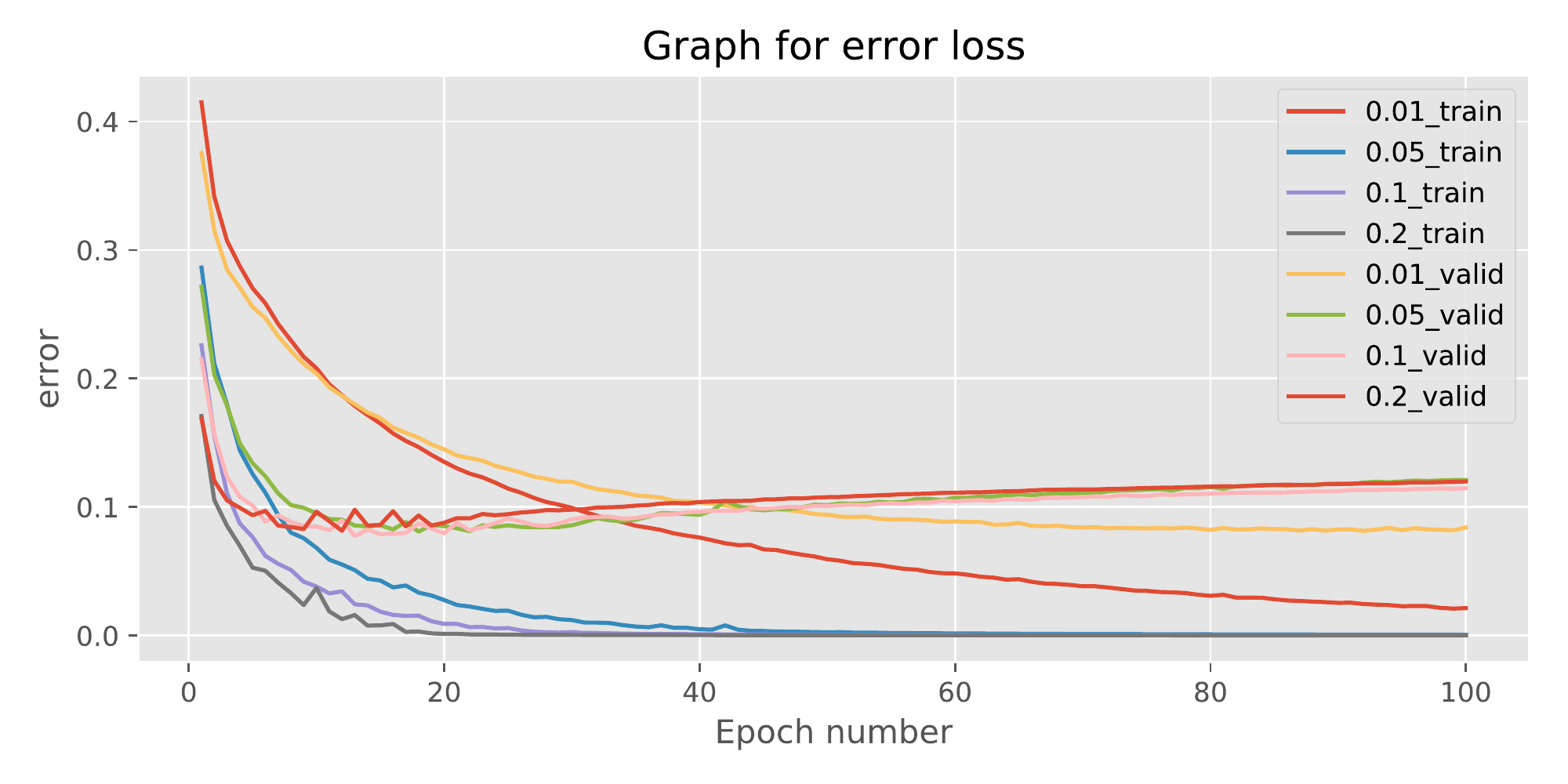}}
\caption{variation of the ELU's error depending on the learning rate }
\label{fig:graph_err_elu}
\end{center}
\vskip -2mm
\end{figure} 

\begin{figure}[hb]
\vskip 2mm
\begin{center}
\centerline{\includegraphics[width=\columnwidth]{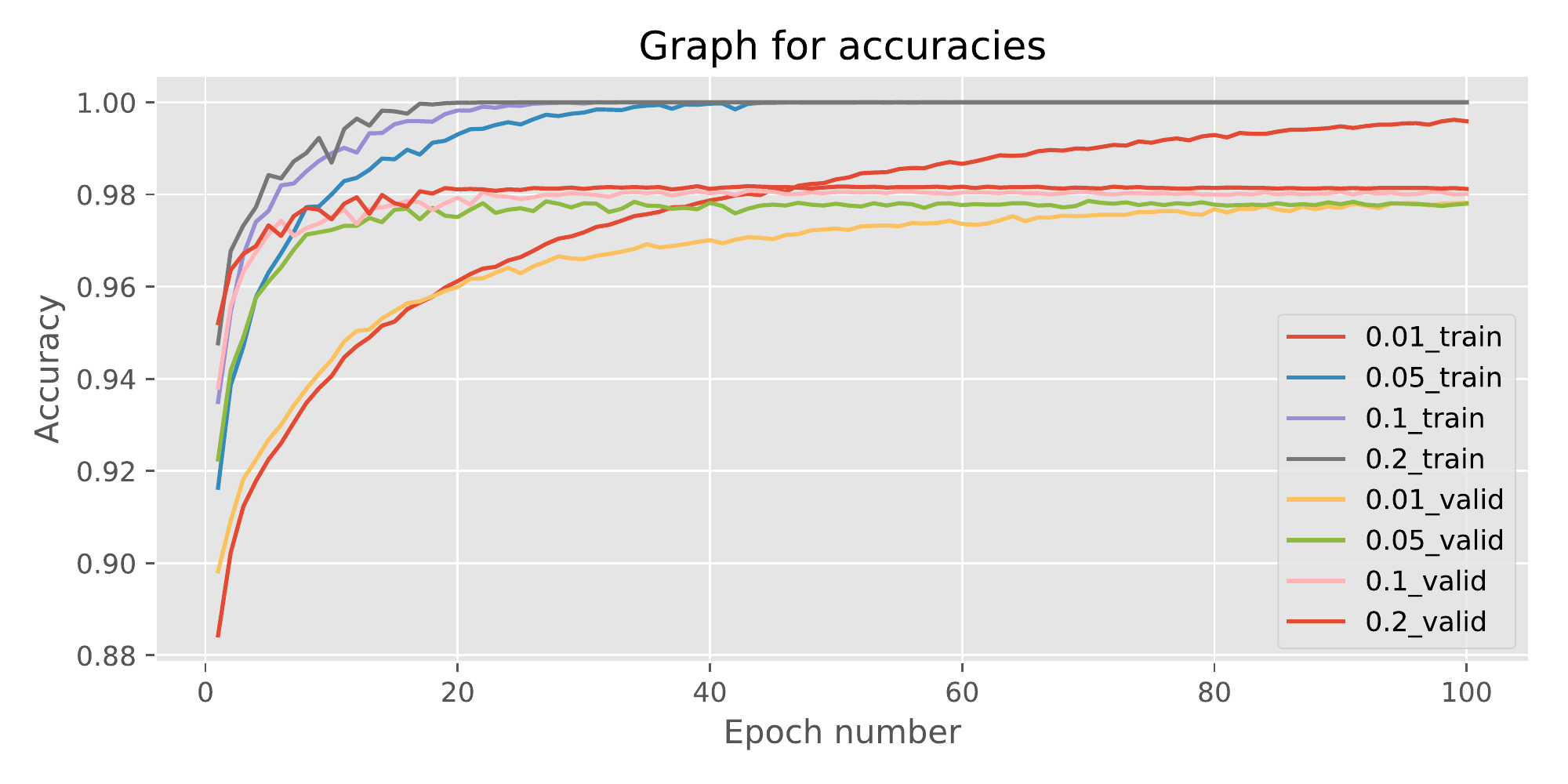}}
\caption{variation of the ELU's accuracies depending on the learning rate }
\label{fig:graph_acc_elu}
\end{center}
\vskip -2mm
\end{figure}
Comparing the different activation functions and looking at the losses, the accuracies and the early overfits behaviour, I have decided to further investigate deeper neural network using the exponential linear unit. 

As it can be observed from the Table 4, ELU performed better than leaky ReLU and ReLU but the the reason why ELU was chosen over SELU is a result of multiple runs. In fact most of the time ELU gave me better results in term of loss, accuracy and how well it generalises compared to SELU. However, it also happened that sometimes SELU activation function was better than ELU, especially when the learning rate was 0.05, but those tests are not reported here because they were done so that I could confirm  that one activation function was clearly and consistently better than the other for the tested ranges of the hyperparameters (learning rate from 0.01 to 0.2, batch size 50 and 100 epochs) but this didn\rq t happen in my case. So I picked the one who performed better on the majority of the cases, however to get a better result it would have been necessary to try and compare all the different combinations of the activation functions with all the parameters which becomes really difficult since the number of possibilities increases rapidly.

\begin{figure}[h]
\vskip 2mm
\begin{center}
\centerline{\includegraphics[width=\columnwidth]{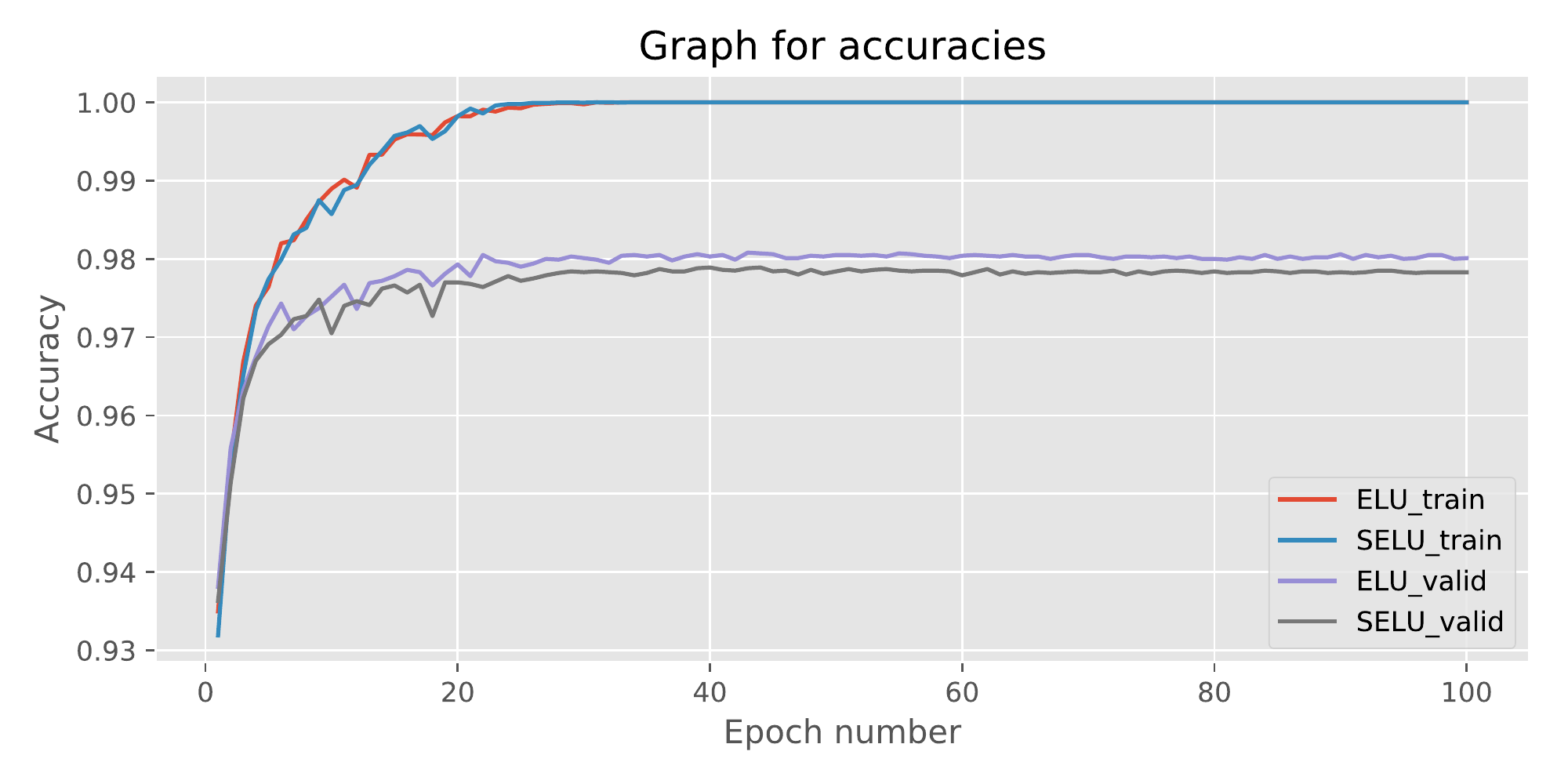}}
\caption{Comparison of accuracies between ELU and SELU}
\label{fig:graph_acc_eluvsselu}
\end{center}
\vskip -2mm
\end{figure}

\begin{figure}[ht]
\vskip 2mm
\begin{center}
\centerline{\includegraphics[width=\columnwidth]{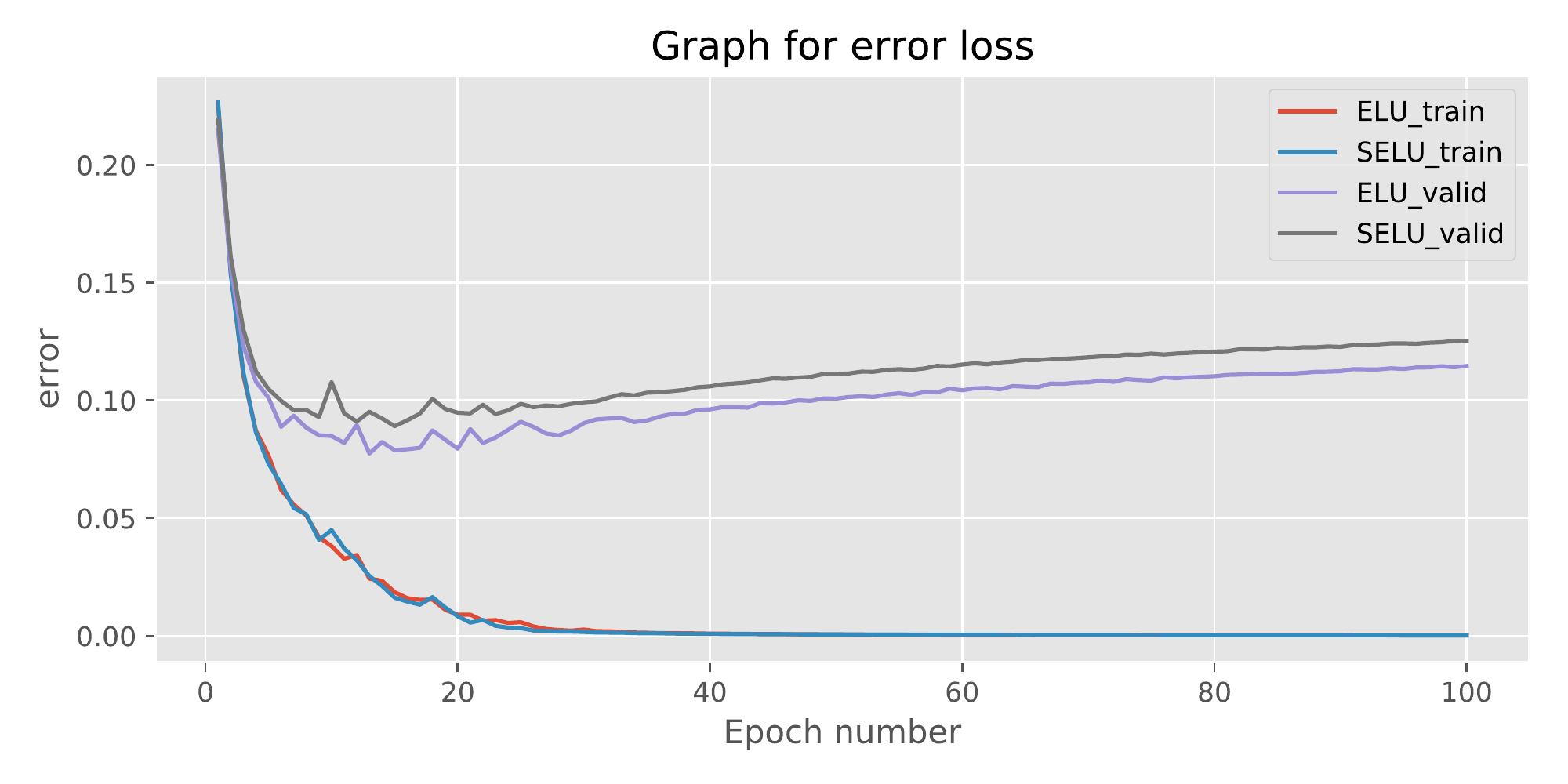}}
\caption{Comparison of error between ELU and SELU}
\label{fig:graph_err_eluvsselu}
\end{center}
\vskip -2mm
\end{figure}
Comparing the ELU with SELU (Figures 3 and 4), the first activation function has 2\% better accuracy than the second one and the loss stays always lower. It can be noticed that around 20 epoch they both start overfitting but ELU starts slightly later then SELU.

Moreover it can be noticed that the accuracies of ReLU and its variants (Table 4) are always higher than the Sigmoid activation function which confirms what has been discusses on \citep{relu00}.

\section{Deep neural network experiments}
\label{sec:dnnexpts}
The next set of experiment focuses on the the effect of using deeper layers with ELU and how this affects the accuracies and the losses particularly on the validation set (Tables 5 and 6).
Before running deeper neural networks, early stopping is executed by making the number of epochs half because it can be seen that with 100 epochs the performance doesn\rq t improve and the loss instead increases. 
For the case when the weights are initialised using Glorot uniform distribution \citep{pmlr-v9-glorot10a}, deeper layers increase the accuracies until 7 hidden layers where the highest accuracy of 0.983 is achieved.
The loss instead tends to alternate between 1.00e-01 and 1.26e-01 by increasing and decreasing as the numbers of hidden layers increase. 

\begin{table}[ht]
\vskip 1mm
\begin{center}
\begin{small}
\begin{sc}
\begin{tabular}{lccccr}
\hline
\abovespace\belowspace
Hidden Layers\.&uniform&f\_in&f\_out&gaussian\\
\hline
\abovespace
2 &    $1.00e_-1$ & $1.07e_-1$ & $1.24e_-1$ & $1.17e_-1$ \\
3 &    $1.09e_-1$ & $1.17e_-1$ & $1.30e_-1$ & $1.21e_-1$\\
4    & $1.26e_-1$ & $1.27e_-1$ & $1.19e_-1$ & $1.26e_-1$ \\
5    & $1.08e_-1$ & $1.10e_-1$ & $1.25e_-1$ & $1.17e_-1$ \\
6    & $1.23e_-1$ & $1.19e_-1$ & $1.45e_-1$ & $1.13e_-1$ \\
7    & $1.16e_-1$ & $1.35e_-1$ & $1.30e_-1$ & $1.14e_-1$\\
8    & $1.23e_-1$ & $1.22e_-1$ & $1.33e_-1$ & $1.19e_-1$ \\
\hline
\end{tabular}
\end{sc}
\end{small}
\caption{Impact on validation errors of different weights initialization methods using ELU as the number of hidden layer increases.}
\label{tab:table-elu-deeplayers}
\end{center}
\vskip -1mm
\end{table}

\begin{table}[ht]
\vskip 1mm
\begin{center}
\begin{small}
\begin{sc}
\begin{tabular}{lccccr}
\hline
\abovespace\belowspace
Hidden Layers\.&uniform&f\_in&f\_out&gaussian\\
\hline
\abovespace
2 &    $0.980$ & $0.979$ & $0.978$ & $0.978$ \\
3 &    $0.981$ & $0.980$ & $0.979$ & $0.980$\\
4    & $0.980$ & $0.981$ & $0.981$ & $0.981$\\
5    & $0.982$ & $0.983$ & $0.982$ & $0.982$\\
6    & $0.982$ & $0.982$ & $0.981$ & $0.981$\\
7    & $0.983$ & $0.982$ & $0.983$ & $0.983$\\
8    & $0.982$ & $0.982$ & $0.982$ & $0.982$\\
\hline
\end{tabular}
\end{sc}
\end{small}
\caption{Impact on validation accuracies of different weights initialization methods on ELU as the number of hidden layer increases.}
\label{tab:table-elu-deeplayers_acc}
\end{center}
\vskip -1mm
\end{table}

When the weights are initialised using fan\_in then the accuracy on the validation set increases more rapidly reaching the highest of 0.983 when 5 hidden layers are used. After this, deeper layers don\rq t improve anymore the accuracy which stays stable at 0.982.

Using fan\_out gives similar results as Glorot uniform distribution regarding the accuracies but it performs worse for the losses whose values, on average, increase.
In particular the loss becomes really small on the training set compared to other initializers.

The accuracy increases with deeper layers also when the weights are drawn from a Glorot gaussian distribution \citep{pmlr-v9-glorot10a}.

\begin{figure}[h]
\vskip 2mm
\centering{
\subfigure[validation error]{
\includegraphics[width=\columnwidth]{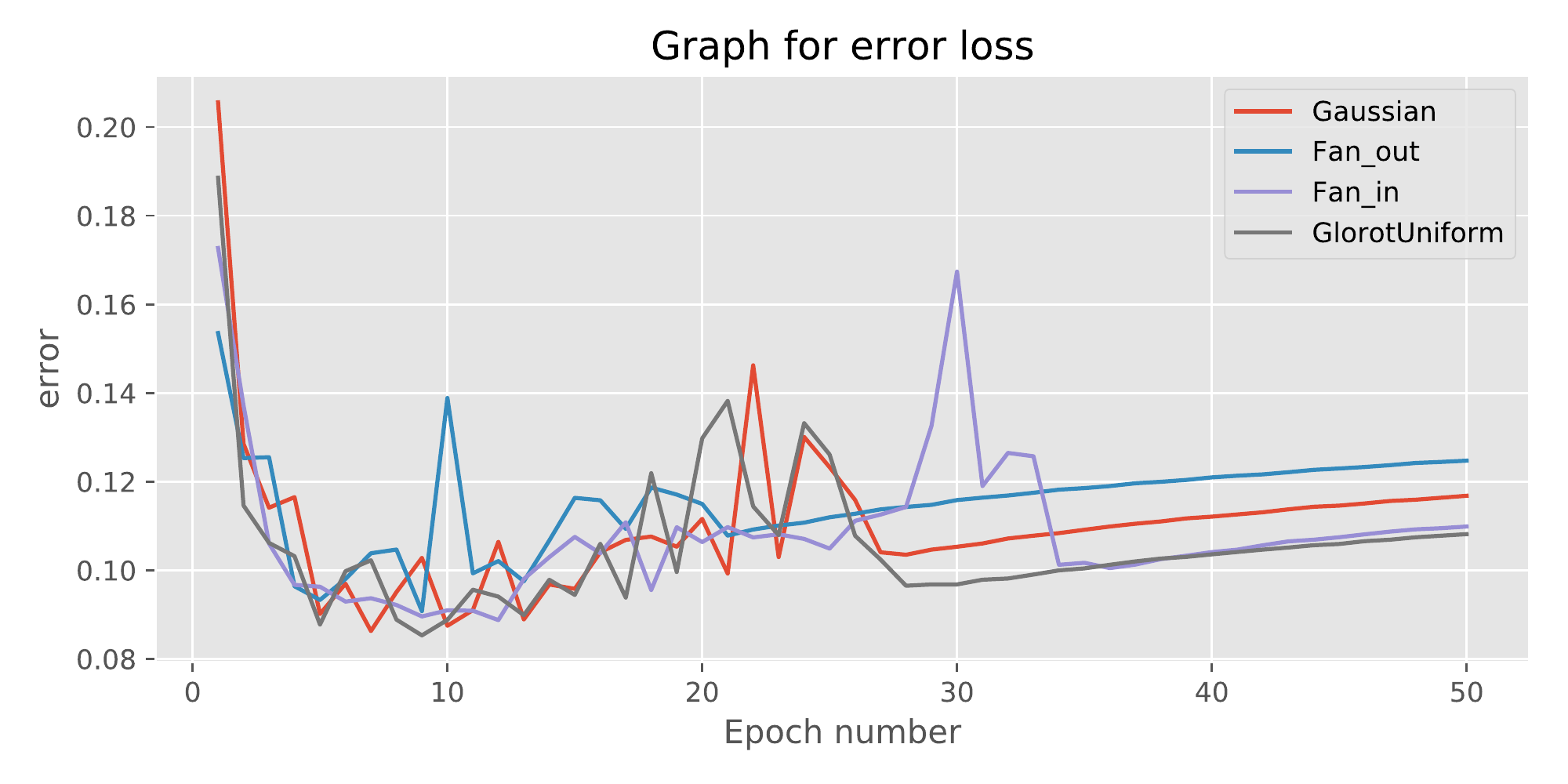}}
\label{fig:graph_err_elu}
\subfigure[validation accuracy]{
\includegraphics[width=\columnwidth]{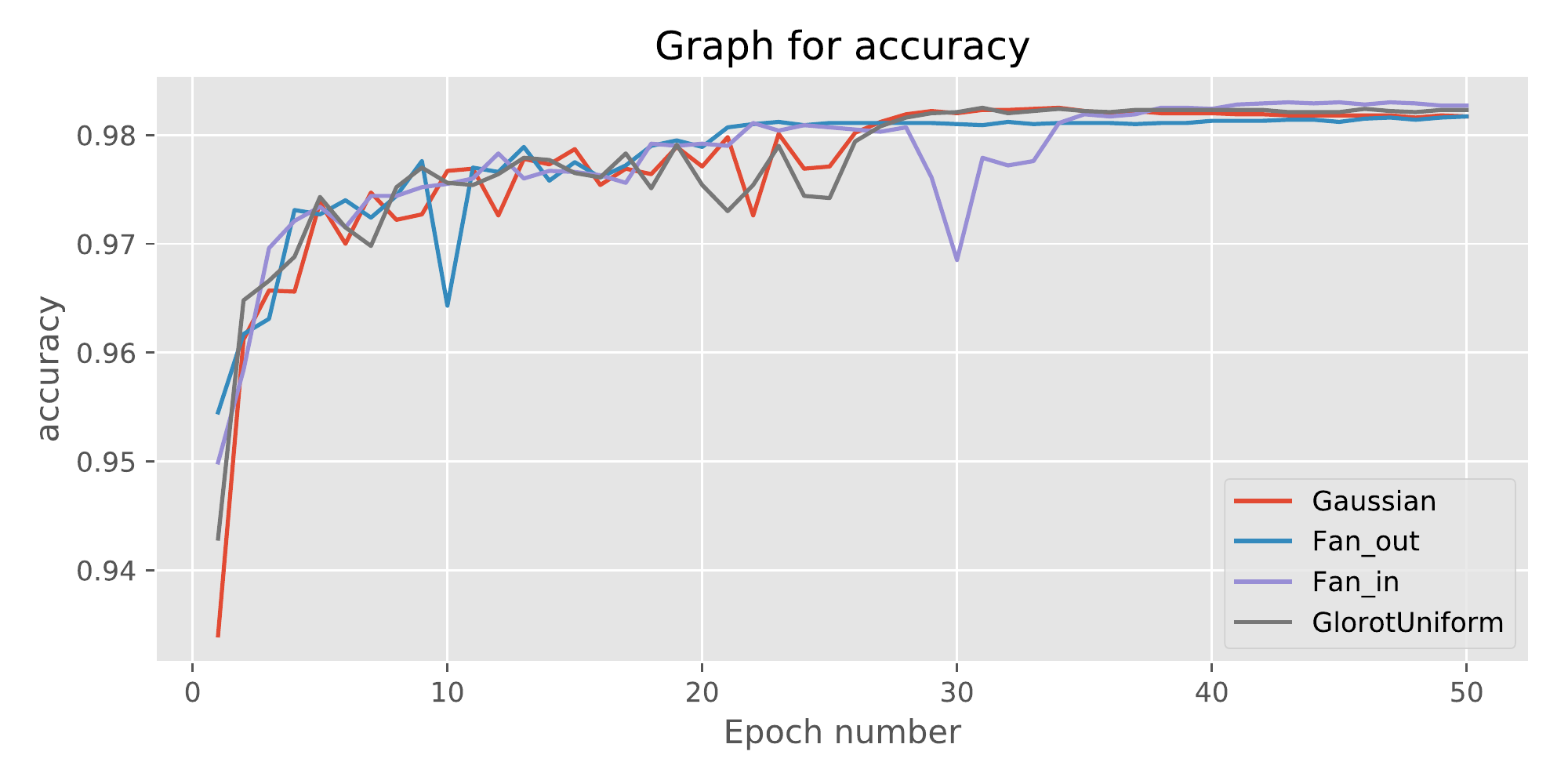}}
\label{fig:graph_err_elu}
}
\caption{ELU's different weights initialization with 5 hidden layers}
\vskip -2mm
\end{figure}

With all four weight initialisation method the accuracies increase from shallower layers to deeper layers. In particular the Glorot uniform distribution gives better results on average as it starts with accuracy of 0.980 when there are 2 hidden layers compared to the others initialisation which  start with an accuracy of 0.978. This in particular can be seen on Figure 5 which focuses on an architecture with five hidden layers.

Moreover I have investigated the effect of deeper layers with SELU with learning rate of 0.05 and compared the results when the weights are initialised with Gaussian distribution with mean 0 and variance $1/n_i$ where $n_i$ is the number of incoming connections \citep{selu00} and Uniform distribution \citep{pmlr-v9-glorot10a}.
\begin{table}[h]
\vskip 1mm
\begin{center}
\begin{small}
\begin{sc}
\begin{tabular}{lccccr}
\hline
\abovespace\belowspace
Hidden Layers\.&U(err)&gauss(err)&U(acc)&gauss(acc)\\
\hline
\abovespace
2 &    $1.01e_-1$ & $1.04e_-1$ & $0.977$ & $0.976$ \\
3 &    $1.11e_-1$ & $1.08e_-1$ & $0.979$ & $0.979$\\
4    & $1.16e_-1$ & $1.17e_-1$ & $0.978$ & $0.980$ \\
5    & $1.19e_-1$ & $1.18e_-1$ & $0.980$ & $0.980$ \\
6    & $1.21e_-1$ & $1.18e_-1$ & $0.981$ & $0.980$ \\
7    & $1.20e_-1$ & $1.16e_-1$ & $0.981$ & $0.982$\\
8    & $1.21e_-1$ & $1.12e_-1$ & $0.981$ & $0.981$ \\
\hline
\end{tabular}
\end{sc}
\end{small}
\caption{Impact on validation errors and accuracies of Glorot uniform vs SELU gaussian weights initialisation using SELU, as the number of hidden layer increases.}
\label{tab:table-elu-deeplayers_acc}
\end{center}
\vskip -2mm
\end{table}

\begin{figure}[h]
\vskip 2mm
\centering{
\subfigure[validation error]{
\includegraphics[width=\columnwidth]{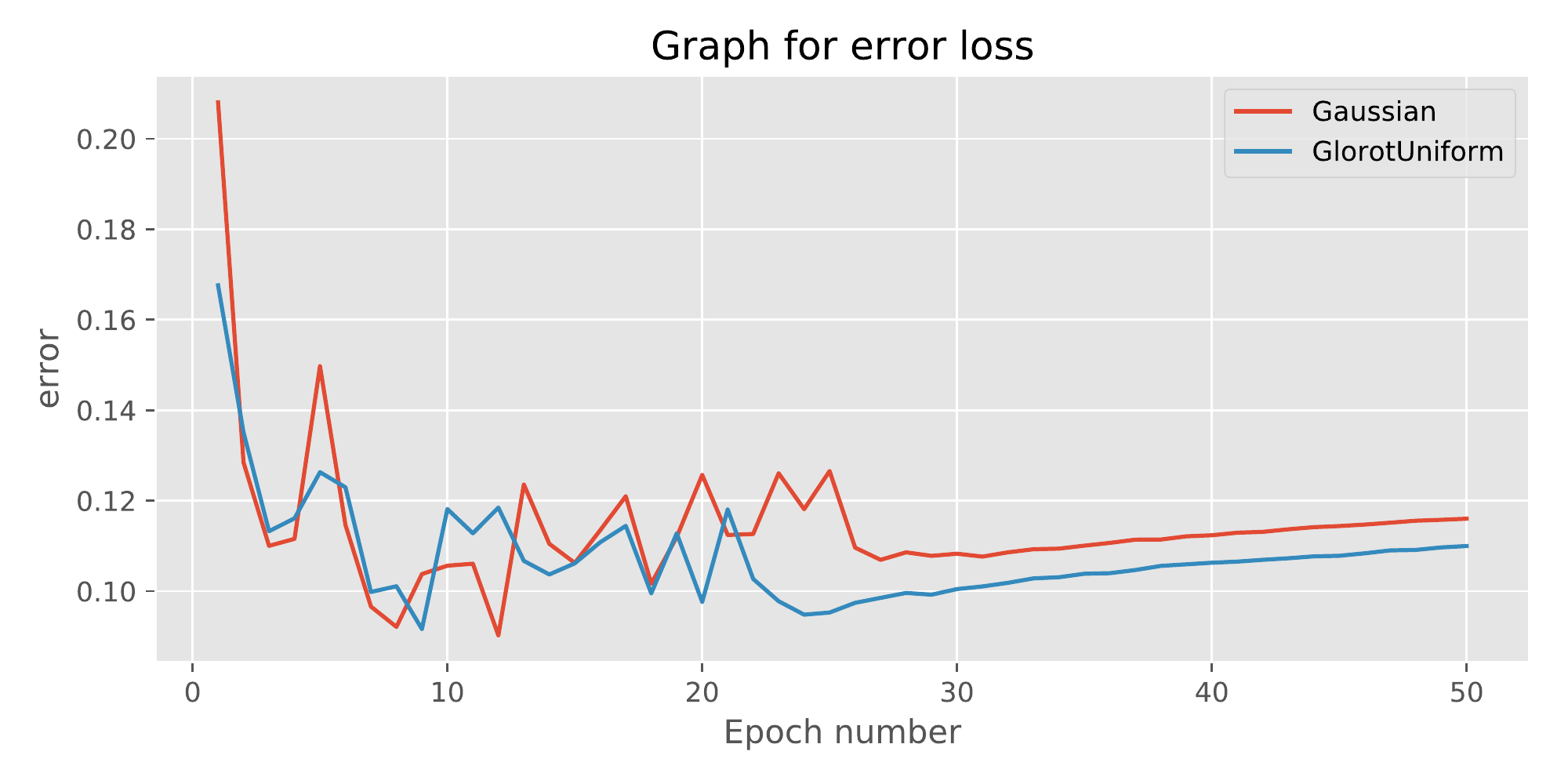}}
\label{fig:graph_err_elu}
\subfigure[validation accuracy]{
\includegraphics[width=\columnwidth]{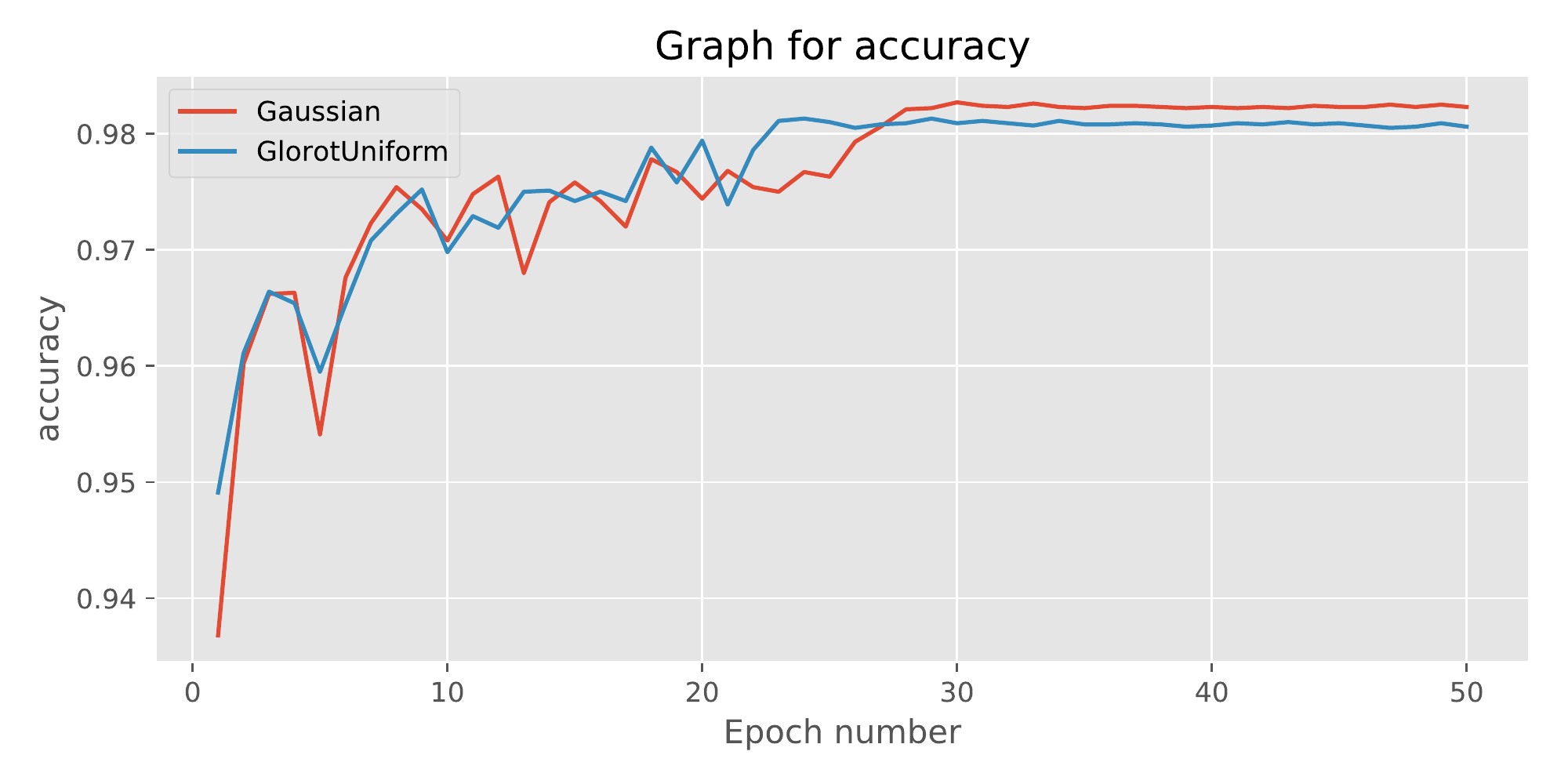}}
\label{fig:graph_err_elu}
}
\caption{SELU's gaussian distribution vs uniform distribution weights initialization with 7 hidden layers}
\vskip -2mm
\end{figure} 

It can be seen from table 7 that in both of them the accuracy increases as the layers are deeper but Gaussian distribution gives better results in term of accuracies. Regarding the losses, the Gaussian distribution gives more stable values of error compared to the other one where the values oscillate more. The better performance of Gaussian distribution over the uniform one is highlighted on Figure 6 which shows the error and accuracy distribution on a model with seven hidden layers.

I have particularly focused on the validation set because the training set gave the same final accuracy of 1, as expected, for all the weights initialisation methods. Similar happens with the errors which become really small as the network becomes deeper. 

After running all those experiments I can say that the weight initialisation methods indeed affect the final accuracies and the losses. As the network becomes deeper, the performance improves as well and the model becomes more sensitive to the weights initialisation. However, if in one hand the deeper layers give better accuracy, on the other hand, they also increase the time required to train hence their computation becomes an issue. As discussed earlier when an activation function has to be chosen, motivations depend on which aspect we are more interested into and it is necessary to find always a good trade-off between pros and cons and, in this case, accuracy vs computation.

Moreover the results follow what has been discussed on \citep{pmlr-v9-glorot10a} where it is discussed the advantage of initialising the weight on non linear function such as  Sigmoid and Tanh. Here the same happens with the ELU activation function where Glorot uniform initialisation performed the best compared to the others.

\section{Conclusions}
\label{sec:concl}
After introducing the ReLU activation function and its variants such as Leaky ReLU, ELU and SELU, the performance between them has been discussed. Furthermore, the effect of deeper neural network had been analysed and the consequences of different weight initialisation methods such as Glorot uniform, fan\_in, fan\_out, Glorot gaussian and SELU gaussian \citep{selu00}.
The experiment can be considered concluded with positive results because it highlights the discussion \citep{pmlr-v9-glorot10a} regarding the weight initialisation and \citep{elu00} where the advantages of ELU over ReLU are stated.
The learning rate is faster in ELU and SELU compared to the ReLU and Leaky ReLU and this confirms the discussion on \citep{elu00} and \citep{selu00}.

It would also be interesting to investigate the effect of the different error functions on the final performance accuracies and relate them with the finding of the ReLU, Leaky ReLU, ELU, SELU and \citep{pmlr-v9-glorot10a} papers.

\bibliography{example-refs}

\begin{thebibliography}{6}
\providecommand{\natexlab}[1]{#1}
\providecommand{\url}[1]{\texttt{#1}}
\expandafter\ifx\csname urlstyle\endcsname\relax
  \providecommand{\doi}[1]{doi: #1}\else
  \providecommand{\doi}{doi: \begingroup \urlstyle{rm}\Url}\fi

\bibitem[Clevert et~al.(2015)Clevert, Unterthiner, and Hochreiter]{elu00}
Clevert, Djork{-}Arn{\'{e}}, Unterthiner, Thomas, and Hochreiter, Sepp.
\newblock Fast and accurate deep network learning by exponential linear units
  (\uppercase{ELU}s).
\newblock arXiv preprint arXiv:1511.07289, 2015.

\bibitem[Glorot \& Bengio(2010)Glorot and Bengio]{pmlr-v9-glorot10a}
Glorot, Xavier and Bengio, Yoshua.
\newblock Understanding the difficulty of training deep feedforward neural
  networks.
\newblock In Teh, Yee~Whye and Titterington, Mike (eds.), \emph{Proceedings of
  the Thirteenth International Conference on Artificial Intelligence and
  Statistics}, volume~9 of \emph{Proceedings of Machine Learning Research},
  pp.\  249--256. PMLR, 13--15 May 2010.

\bibitem[Klambauer et~al.(2017)Klambauer, Unterthiner, Mayr, and
  Hochreiter]{selu00}
Klambauer, G{\"{u}}nter, Unterthiner, Thomas, Mayr, Andreas, and Hochreiter,
  Sepp.
\newblock Self-normalizing neural networks.
\newblock arXiv preprint arXiv:1706.02515, 2017.

\bibitem[LeCun(1999)]{mnist00}
LeCun, Yann.
\newblock The \uppercase{MNIST D}ataset \uppercase{O}f \uppercase{H}andwritten
  \uppercase{D}igits (\uppercase{I}mages).
\newblock 1999.

\bibitem[Maas et~al.(2013)Maas, Hannun, and Ng]{lrelu00}
Maas, Andrew~L, Hannun, Awni~Y, and Ng, Andrew~Y.
\newblock Rectifier nonlinearities improve neural network acoustic models.
\newblock volume~30. In Proc. ICML, 2013.

\bibitem[Nair \& Hinton(2010)Nair and Hinton]{relu00}
Nair, Vinod and Hinton, Geoffrey~E.
\newblock Rectified linear units improve restricted boltzmann machines.
\newblock pp.\  807--€"814. In Proc. ICML, volume 30, 2010.

\end{thebibliography}

\end{document}